\PassOptionsToPackage{unicode}{hyperref}
\PassOptionsToPackage{hyphens}{url}
\PassOptionsToPackage{dvipsnames,svgnames,x11names}{xcolor}
\documentclass[
  aps,
  pre,
  reprint,
  superscriptaddress,
  amsmath,
  amssymb,
  floatfix,
  longbibliography]{revtex4-2}
\usepackage{xcolor}
\usepackage{amsmath,amssymb}
\usepackage{iftex}
\ifPDFTeX
  \usepackage[T1]{fontenc}
  \usepackage[utf8]{inputenc}
  \usepackage{textcomp} 
\else 
  \usepackage{unicode-math} 
  \defaultfontfeatures{Scale=MatchLowercase}
  \defaultfontfeatures[\rmfamily]{Ligatures=TeX,Scale=1}
\fi
\usepackage{lmodern}
\ifPDFTeX\else
\fi
\IfFileExists{upquote.sty}{\usepackage{upquote}}{}
\IfFileExists{microtype.sty}{
  \usepackage[]{microtype}
  \UseMicrotypeSet[protrusion]{basicmath} 
}{}
\usepackage{graphicx}
\makeatletter
\newsavebox\pandoc@box
\newcommand*\pandocbounded[1]{
  \sbox\pandoc@box{#1}%
  \Gscale@div\@tempa{\textheight}{\dimexpr\ht\pandoc@box+\dp\pandoc@box\relax}%
  \Gscale@div\@tempb{\linewidth}{\wd\pandoc@box}%
  \ifdim\@tempb\p@<\@tempa\p@\let\@tempa\@tempb\fi
  \ifdim\@tempa\p@<\p@\scalebox{\@tempa}{\usebox\pandoc@box}%
  \else\usebox{\pandoc@box}%
  \fi%
}
\def\fps@figure{htbp}
\makeatother
\providecommand{\tightlist}{%
  \setlength{\itemsep}{0pt}\setlength{\parskip}{0pt}}
\makeatletter
\@ifpackageloaded{caption}{}{\usepackage{caption}}
\AtBeginDocument{%
\ifdefined\contentsname
  \renewcommand*\contentsname{Table of contents}
\else
  \newcommand\contentsname{Table of contents}
\fi
\ifdefined\listfigurename
  \renewcommand*\listfigurename{List of Figures}
\else
  \newcommand\listfigurename{List of Figures}
\fi
\ifdefined\listtablename
  \renewcommand*\listtablename{List of Tables}
\else
  \newcommand\listtablename{List of Tables}
\fi
\ifdefined\figurename
  \renewcommand*\figurename{Figure}
\else
  \newcommand\figurename{Figure}
\fi
\ifdefined\tablename
  \renewcommand*\tablename{Table}
\else
  \newcommand\tablename{Table}
\fi
}
\@ifpackageloaded{float}{}{\usepackage{float}}
\floatstyle{ruled}
\@ifundefined{c@chapter}{\newfloat{codelisting}{h}{lop}}{\newfloat{codelisting}{h}{lop}[chapter]}
\floatname{codelisting}{Listing}

\makeatother
\makeatletter
\@ifpackageloaded{caption}{}{\usepackage{caption}}
\@ifpackageloaded{subcaption}{}{\usepackage{subcaption}}
\makeatother
\usepackage{bookmark}
\IfFileExists{xurl.sty}{\usepackage{xurl}}{} 
\hypersetup{
  pdftitle={Latent Spectroscopy: Posterior Collapse as a Feature},
  pdfauthor={Johannes Hirn},
  colorlinks=true,
  linkcolor={blue},
  filecolor={Maroon},
  citecolor={Blue},
  urlcolor={Blue},
  pdfcreator={LaTeX via pandoc}}

\begin{document}

\title{Latent Spectroscopy: Posterior Collapse as a Feature}

\author{Johannes Hirn}

\affiliation{Image Processing Laboratory (IPL), Universitat de València,
Paterna, València 46980, Spain}

\email{hirnjo@uv.es}

\homepage{https://orcid.org/0000-0003-0267-2479}

\begin{abstract}
We show that, in linear Gaussian VAEs, posterior collapse is a form of
latent feature selection.

Feature importance is set by each latent coordinate's contribution to
reconstruction, itself given by the corresponding PCA eigenvalue.

Varying the regularizer strength \(\beta\) reveals a ranked spectrum of
collapse events, with thresholds set by the utility/PCA spectrum.

A mode-by-mode analysis identifies the scale-invariant signal fraction
as an order parameter obeying Landau scaling near collapse.

WorldClim experiments confirm both the utility-threshold calibration and
the predicted near-collapse scaling.
\end{abstract}

\maketitle

\section{Introduction}\label{introduction}

Variational autoencoders (VAEs) learn compressed probabilistic
representations by balancing data fidelity against the divergence from a
reference latent prior
\citep{kingma2014autoencoding, kingma2019introvae}. \(\beta\)-VAEs make
this rate--distortion tradeoff explicit by introducing the
regularization strength \(\beta\) as a control parameter multiplying the
relative entropy term
\citep{higgins2017betavae, burgess2018understandingbetavae}.

In practical applications, it has often been observed that many latent
coordinates revert to the prior, stop carrying meaningful information
about the input, and are effectively ignored by the decoder. Much of the
VAE literature treats this posterior collapse as a failure mode,
attributing it to decoder expressivity, bad local minima, amortization
lag, or a mismatch between optimization dynamics and the variational
objective; many approaches have therefore been proposed to delay or
soften it
\citep{bowman2016generating, alemi2018fixingbrokenelbo, he2019lagging, dai2020usual, rybkin2021calibrateddecoders}.
Even in the linear case, collapse is usually considered a pathology
\citep{lucas2019dontblame, ichikawa2024learningdynamics, bai2025highdimvae}.

Important prior work already softens this pathology-only view. Dai and
coauthors showed that VAEs can deliberately ignore nuisance or outlier
structure, and later argued that selective collapse can be benign or
even necessary for good sampling
\citep{dai2017hiddentalents, dai2019diagnosing, dai2020usual}. Lucas et
al.~and Dai--Wipf further show that collapse can arise from the
objective landscape itself, including decoder-noise choices, large KL
weights, and local extrema, not only from optimizer malfunction
\citep{lucas2019dontblame, dai2020usual}. Optimization effects can of
course modify access to that equilibrium: learning-rate schedules,
annealing protocols, early stopping, and decoder parameterization can
make collapse more severe, delay it, or obscure the equilibrium regime
selected by the loss.

Yet, posterior collapse is not inherently good or bad: whether one seeks
to avoid, delay, or exploit it depends on the downstream goal. When the
latent representation itself is the object of interest, as in
dimensionality reduction, feature learning, and the interpretability of
effective variables, posterior collapse can be a feature rather than a
bug.

Here we study posterior collapse as a measurable response to the
regularization strength, focusing on the linear Gaussian case as a
baseline. Rate--distortion analyses of VAEs, including the
information-theoretic ELBO view of Alemi et al., already emphasize the
global tradeoff between total distortion \(D\) and total rate \(R\)
\citep{alemi2018fixingbrokenelbo}. At that level, changing the
regularization strength \(\beta\) moves the model along a smooth
\(D(R)\) curve.

Our refinement is to ask whether that smooth curve can be resolved into
coordinate-level events. Rather than asking only where the model lies in
the \((D,R)\) plane, we ask whether one can identify which latent
coordinates remain active at each point along the curve, how much
distortion each coordinate removes, and whether their disappearance
occurs through a reproducible sequence of thresholds rather than through
collective fading. If the latent representation is truncated to its
first \(k-1\) ordered coordinates, the marginal reduction in distortion
obtained by adding the \(k\)-th coordinate defines its reconstruction
utility. In this reconstruction task, that utility is the
feature-importance score of the ranked latent feature.

We refer to this logic as latent spectroscopy: probing a learned
representation by varying a control parameter and reading off
latent-coordinate spectra.

This perspective also places VAEs in contact with the information
bottleneck (IB) program. The original IB formulation of Tishby, Pereira,
and Bialek casts representation learning as a competition between
compression and information \citep{tishby2000informationbottleneck}. The
Gaussian IB solution of Chechik, Globerson, Tishby, and Weiss makes the
spectral structure explicit: different bottleneck dimensions appear
sequentially through a cascade controlled by the spectrum of a quadratic
operator \citep{chechik2005gaussianib}.

The central claim established below is that, in the linear Gaussian VAE,
this spectral picture is exact: the collapse spectrum, utility spectrum,
and normalized PCA spectrum coincide component by component. The linear
case avoids basis rotations induced by nonlinear feature learning and
suppresses architecture-dependent mixing, while preserving the essential
rate--distortion competition.

The connection between linear Gaussian VAEs and PCA is well known. The
goal here is not to rediscover PCA as an endpoint, but to show how the
variational objective reaches that endpoint component by component. When
scanned toward stronger regularization, we find a sequence of collapses;
scanned in the opposite direction, the same sequence corresponds to PCA
components entering one by one, thus improving the reconstruction.

Normalization conventions matter: raw \(\beta\) values are not
universal, but depend on input scaling, loss normalization, and
reconstruction convention. Once \(\beta\) is compared with the data
variance measured in decoder-variance units, however, collapse
thresholds and utilities can be compared directly with normalized PCA
eigenvalues (explained-variance ratios).

All empirical figures in the present paper are for linear VAEs trained
on WorldClim \citep{fick2017worldclim2}: a gridded global climatology of
19 bioclimatic variables at 10 arc-minute resolution. We use
geography-aware train/validation/test splits to reduce spatial leakage.
The final figures were computed on the held-out test split after the
analysis protocol was fixed.

The paper is organized as follows. Section II defines the scan protocol:
normalized scan units, latent order parameters, collapse spectra,
utility spectra, and the signal-fraction ranking convention. Section III
gives the single-mode Landau derivation and identifies the
scale-invariant signal fraction as the order parameter. Section IV shows
that, in the linear Gaussian case, the collapse spectrum, utility
spectrum, and PCA spectrum are the same object. Section V interprets the
result as a spectral cutoff exactly calibrated by marginal utility.
Section VI concludes by explaining what this linear baseline
establishes. Section VII outlines the broader latent spectroscopy
program beyond the linear Gaussian case.

\section{Collapse Spectroscopy}\label{collapse-spectroscopy}

The loss used throughout is the \(\beta\)-VAE objective

\begin{equation}\label{eq-vae-loss}
\mathcal L
=D+\beta R,
\end{equation}

where the reconstruction distortion is the expected negative
log-likelihood, \begin{equation}\label{eq-auto-001}
D =
-\overline{
\left\langle
\log p_\theta(x\mid z)
\right\rangle
},
\end{equation}

and the latent regularization term is the posterior average of the
log-density ratio to the normal prior \(p(z)=\mathcal N(0,I)\),
\begin{equation}\label{eq-auto-002}
R =
\overline{
\mathrm{KL}\!\left[q_\phi(z\mid x)\,\|\,p(z)\right]
} =
\overline{
\left\langle
\log q_\phi(z\mid x)-\log p(z)
\right\rangle
}
.
\end{equation}

Throughout this paper, overbars denote averages over the data
distribution; in particular, \(\bar x\) denotes the data mean. Angled
brackets denote posterior averages over latent noise conditional on a
data point: \(\langle f(z,x)\rangle\) is the posterior expectation of
\(f\) at fixed \(x\) under the posterior.

In the experiments and derivations below, the decoder likelihood is
Gaussian with fixed isotropic variance,
\begin{equation}\label{eq-auto-003}
p_\theta(x\mid z)
=\mathcal N\!\left(\hat x_\theta(z),\sigma_\text{dec}^2 I\right).
\end{equation}

For this fixed-variance Gaussian likelihood, the distortion can be
rewritten (up to an additive constant) as a sum of squares
\begin{equation}\label{eq-auto-004}
D =
\frac{1}{2\sigma_\text{dec}^2}
\overline{
\left\langle
\|x-\hat x_\theta(z)\|^2
\right\rangle
}.
\end{equation}

We now introduce our normalizations, using the total data variance
\begin{equation}\label{eq-auto-005}
V \equiv \overline{\|x-\bar x\|^2}
=\sum_k \lambda_k,
\end{equation}

where \(\lambda_k\) are the PCA eigenvalues of the centered data. We
report squared reconstruction errors in units of this variance,
\begin{equation}\label{eq-auto-006}
\tilde D
\equiv
\frac{
\overline{
\left\langle
\|x-\hat x_\theta(z)\|^2
\right\rangle
}
}{V},
\end{equation}

which implies that the optimized fully collapsed reconstruction has unit
normalized distortion \begin{equation*}
\left. \tilde D \right|_{q_\phi(z \mid x) = p(z) } = 1.
\end{equation*} Pulling out a common prefactor, we recast the original
loss in a way that makes explicit the real tradeoff between normalized
distortion and information \begin{equation}\label{eq-normalized-loss}
\mathcal L =
\frac{V}{2\sigma_\text{dec}^2}
\left(\tilde D+2TR\right),
\end{equation}

Raw \(\beta\) is not meaningful by itself: it must be compared to the
total data variance measured in decoder-variance units,
\(V/\sigma_\text{dec}^2\). We therefore introduce a new notation for the
effective control parameter \begin{equation}\label{eq-scan-temperature}
T
\equiv
\frac{\beta}{V/\sigma_\text{dec}^2}
=
\frac{\beta\sigma_\text{dec}^2}{V}.
\end{equation}

Hereafter, we refer to \(T\) as an effective temperature; Appendix
\ref{sec-relative-entropy-effective-temperature} elaborates on the
meaning of this terminology. For the main text, the important point is
simpler: \(T\) controls the balance between the ordered, input-dependent
latent state and the disordered prior state.

Note that the control parameter \(T\) has several interpretations. It
was initially introduced as a normalized version of the regularization
strength \(\beta\). In rate--distortion language, it is the price of one
nat of latent information, measured in normalized distortion units. In
statistical-physics language, it is an effective temperature. In the
language used below, it is a spectral cutoff: latent coordinates
collapse one by one as \(T\) is raised. In the linear Gaussian case this
cutoff is exactly calibrated by marginal reconstruction utility.

Together, Eqs. \eqref{eq-scan-temperature} and
\eqref{eq-normalized-loss} show that, in this fixed-variance Gaussian
setting, \(\beta\) and \(\sigma_\text{dec}^2\) are not independent
control axes: only the product \(\beta\sigma_\text{dec}^2\) controls the
balance between the quadratic reconstruction term and the rate.

This normalization is important in practice. Statements such as
``\(\beta=1\) gives \(K\) active latent dimensions'' are not intrinsic
properties of the data or the model: they depend on input scaling, loss
normalization, and reconstruction convention. By using normalized units,
one obtains a control parameter \(T\) that can be meaningfully compared
across datasets. For instance, in the extreme case where a single
spectral mode carries essentially all the variance, that mode would
collapse at \(T \simeq 1\) as we will see below, and in general we have:
\begin{equation*}
\tilde D(T > 1) \simeq 1 .
\end{equation*} For each latent coordinate \(k\), we monitor four
observables on held-out data,
\begin{equation}\label{eq-latent-observables}
\overline{\mu_k(x)^2},\qquad
\overline{\sigma_k(x)^2},\qquad
\overline{\log \sigma_k(x)^2},\qquad
R_k(T)=\overline{\mathrm{KL}_k(x)},
\end{equation}

with the KL divergence between two Gaussians given by
\begin{equation}\label{eq-explicit-KL}
\begin{aligned}
\mathrm{KL}_k(x)
&=
\frac{1}{2}
\Bigl[
\mu_k(x)^2+\sigma_k(x)^2 \\
&\qquad
-\log\sigma_k(x)^2-1
\Bigr].
\end{aligned}
\end{equation}

For the standard Gaussian prior, the KL identity above relates these
four observables algebraically. The Jensen gap between the log of the
mean variance and the mean log-variance
\begin{equation}\label{eq-auto-047}
J_k(T)
=\log\overline{\sigma_k(x)^2}-\overline{\log\sigma_k(x)^2}.
\end{equation}

is used later as a diagnostic of posterior-variance heterogeneity.

Our default ranking and fitting observable is the posterior signal
fraction \begin{equation}\label{eq-signal-fraction}
M_k^2(T) =
\frac{\overline{\mu_k(x)^2}}
{\overline{\mu_k(x)^2}+\overline{\sigma_k(x)^2}}.
\end{equation}

Equivalently, we can rewrite this signal fraction as
\begin{equation}\label{eq-auto-011}
M_k^2(T)=\frac{\mathrm{SNR}_k(T)}{1+\mathrm{SNR}_k(T)}
\end{equation}

in terms of the signal-to-noise ratio
\begin{equation}\label{eq-auto-012}
\mathrm{SNR}_k(T)
=\frac{\overline{\mu_k(x)^2}}{\overline{\sigma_k(x)^2}}.
\end{equation}

Thus \(M_k^2\) and the posterior SNR induce the same ranking of latent
dimensions. This SNR-like relevance criterion follows the active-unit
heuristics used in some interpretability analyses of VAE representation
learning
\citep{barenboim2024music, sanz2026learningsymmetries, sanz2026artificialsymmetries}.
Section III explains why the ratio \(M_k^2\) \eqref{eq-signal-fraction}
is selected by the one-mode loss.

The collapse threshold \(T_k\) of ranked mode \(k\) is extracted from
the boundary between the active ordered state and the collapsed
disordered state. Read in the opposite scan direction, the same value is
the onset threshold at which the mode becomes active. The primary fit
uses the one-mode law to extract a threshold \(T_k\)
\begin{equation}\label{eq-signal-fraction-fit}
M_k^2(T)=\left[1-\frac{T}{T_k}\right]_+.
\end{equation}

Because the collapse threshold is measured for each ranked latent
coordinate, the result is not a single crossover value of \(\beta\) but
an ordered spectrum \begin{equation}\label{eq-auto-013}
\{T_1,T_2,\ldots\}.
\end{equation}

This collapse-threshold spectrum is the first of the two spectra in the
paper.

The second spectrum is obtained from truncated reconstructions. Let
\(\tilde D_k\) denote the normalized distortion achieved by keeping the
first \(k\) ranked latent coordinates and pruning the rest. Then the
marginal utility of the \(k\)-th coordinate is
\begin{equation}\label{eq-auto-014}
\Delta \tilde D_k = \tilde D_{k-1} - \tilde D_k.
\end{equation}

These utilities define a ranked spectrum of distortion reductions. In a
weaker picture, collapse thresholds and utilities could have been merely
correlated: modes that collapse later might simply also tend to be more
useful.

For the linear Gaussian baseline, the stronger claim derived below is
that the two spectra coincide:
\begin{equation}\label{eq-utility-onset-duality}
T_k = \Delta \tilde D_k.
\end{equation}

We refer to the relation \eqref{eq-utility-onset-duality} as the mode
utility--threshold duality.

\section{Single-Mode Landau
Derivation}\label{single-mode-landau-derivation}

The simplest setting for studying posterior collapse is a linear VAE
restricted to a single scalar data mode. Consider a centered scalar mode
\(x\) with variance \begin{equation}\label{eq-auto-015}
\lambda = \overline{x^2}.
\end{equation}

For a one-mode linear VAE with Gaussian posterior and linear decoder,
\begin{equation}\label{eq-one-mode-family}
\begin{aligned}
q(z\mid x)&=\mathcal N(ax,\sigma_z^2),\\
\hat x&=wz,\\
p(z)&=\mathcal N(0,1).
\end{aligned}
\end{equation}

the one-mode objective can first be written in terms of averaged
posterior observables as
\begin{equation}\label{eq-one-mode-loss-moments}
\begin{aligned}
\mathcal L
&=
\frac{1}{2\sigma_\text{dec}^2}
\overline{
\left\langle
(x-wz)^2
\right\rangle
}\\
&\quad+
\frac{\beta}{2}
\left[
\overline{\mu(x)^2}
+\overline{\sigma_z^2(x)}
-\overline{\log \sigma_z^2(x)}
-1
\right].
\end{aligned}
\end{equation}

For the constant-variance linear parametrization in Eq.
\eqref{eq-one-mode-family}, Eq. \eqref{eq-one-mode-loss-moments} becomes
\begin{equation}\label{eq-one-mode-loss-coefficients}
\begin{aligned}
\mathcal L
&=
\frac{1}{2\sigma_\text{dec}^2}
\left[
(1-wa)^2\lambda+w^2\sigma_z^2
\right]\\
&\quad+
\frac{\beta}{2}
\left[
a^2\lambda+\sigma_z^2-\log \sigma_z^2-1
\right].
\end{aligned}
\end{equation}

This constant-variance step is an ansatz: its accuracy can be tested via
the Jensen gap \(J_k(T)\) plotted in Appendix
\ref{sec-latent-scale-diagnostics}.

The original optimization has three scalar variables,
\((a,\sigma_z^2,w)\). We can minimize in those variables, or in any
equivalent reparametrization. A useful reparametrization is to separate
a scale-invariant signal fraction
\begin{equation}\label{eq-one-mode-signal-fraction}
M^2 =
\frac{\overline{\mu(x)^2}}
{\overline{\mu(x)^2}+\overline{\sigma_z^2(x)}},
\end{equation}

from the absolute posterior scale (second moment of the aggregate
posterior) \begin{equation}\label{eq-one-mode-scale}
A^2=\overline{\mu(x)^2}+\overline{\sigma_z^2(x)}.
\end{equation}

The key step is the decoder minimization. At fixed posterior statistics,
stationarity with respect to \(w\) fixes
\begin{equation}\label{eq-one-mode-decoder}
\begin{aligned}
w_\star
&=
\frac{a\lambda}
{\overline{\mu(x)^2}+\overline{\sigma_z^2(x)}}.
\end{aligned}
\end{equation}

Substituting this decoder back into the reconstruction term yields
\begin{equation}\label{eq-decoder-eliminated-reconstruction}
\begin{aligned}
\frac{1}{2\sigma_\text{dec}^2}
\overline{
\left\langle
(x-w_\star z)^2
\right\rangle
}
&=
\frac{\lambda}{2\sigma_\text{dec}^2}(1-M^2).
\end{aligned}
\end{equation}

Thus, for the constant-variance one-mode solution, eliminating the
decoder gives the reduced objective
\begin{equation}\label{eq-reduced-one-mode-loss}
\begin{aligned}
\mathcal L_\star(M^2,A^2)
&=
\frac{\lambda}{2\sigma_\text{dec}^2}(1-M^2)\\
&\quad+
\frac{\beta}{2}
\left[
A^2-\log(A^2)-\log(1-M^2)-1
\right],
\end{aligned}
\end{equation}

where the star subscript indicates that the decoder has been set to the
value \(w_\star\) that minimizes the loss at fixed encoder statistics.
Dividing by the collapsed one-mode distortion scale
\(D_0=\lambda/(2\sigma_\text{dec}^2)\) gives
\begin{equation}\label{eq-dimensionless-one-mode-loss}
\begin{aligned}
\ell_\star(M^2,A^2)
&\equiv
\frac{\mathcal L_\star}{D_0}\\
&=
(1-M^2)
+\tau
\left[
A^2-\log(A^2)-\log(1-M^2)-1
\right],
\end{aligned}
\end{equation}

with \begin{equation}\label{eq-def-tau}
\tau = \frac{\beta \sigma_\text{dec}^2}{\lambda}.
\end{equation}

The reduced loss in Eq. \eqref{eq-dimensionless-one-mode-loss} is the
starting point for the branch analysis below.

\subsection{Branch analysis}\label{branch-analysis}

The collapsed branch is the solution in which the posterior equals the
prior, \(\overline{\mu(x)^2}=0\), \(\overline{\sigma_z^2(x)}=1\), and
its Gaussian distortion is the collapsed baseline
\begin{equation}\label{eq-auto-016}
D_0 = \frac{\lambda}{2\sigma_\text{dec}^2}.
\end{equation}

The active branch is obtained by minimizing first with respect to the
signal fraction: \begin{equation}\label{eq-m2-stationarity}
\frac{\partial \ell_\star}{\partial M^2}
=
-1
+
\frac{\tau}{1-M^2},
\end{equation}

yielding \begin{equation}\label{eq-one-mode-active-law}
M^2 = 1-\tau
\end{equation}

provided the condition \begin{equation}\label{eq-local-threshold}
\tau < \tau_c \equiv 1,
\end{equation}

is satisfied, i.e.~the reduced temperature is smaller than the collapse
threshold \(\tau_c=1\). Equivalently, in raw \(\beta\) units this
collapse threshold is \(\beta_c=\lambda/\sigma_\text{dec}^2\) by Eq.
\eqref{eq-def-tau}. This threshold separates the active ordered state
from the disordered collapsed state: for \(\tau > 1\), the active branch
ceases to exist and the collapsed branch remains, so \(M^2=0\).

The Landau form is obtained by expanding the loss near \(M^2=0\):
\begin{equation}\label{eq-one-mode-landau-expansion}
\ell_\star(M^2,A^2)
=
\ell_\star(0,A^2)
+
\left(\tau-1\right)M^2
+
\frac{\tau}{2}(M^2)^2
+
O\!\left((M^2)^3\right).
\end{equation}

The coefficient of \(M^2\) in Eq. \eqref{eq-one-mode-landau-expansion}
changes sign at the collapse point in Eq. \eqref{eq-local-threshold},
and the exact minimizer gives the linear law
\eqref{eq-one-mode-active-law} for \(M^2\). The data enter through
coefficients such as \(\lambda\), which set thresholds and utilities,
while the model class fixes the local onset shape.

Note that the onset shape familiar from Landau theories of continuous
transitions follows here directly from the competition between the
linear reconstruction gain and the exact logarithmic Gaussian KL term,
after the decoder has been minimized. It is not introduced through a
phenomenological free-energy ansatz. In field theory, by contrast, a
Landau functional is typically an effective description of many
interacting degrees of freedom, often obtained after coarse-graining. As
discussed in Appendix \ref{sec-relative-entropy-effective-temperature},
the formal resemblance does not imply an underlying coarse-grained
many-body construction or a thermodynamic large-system limit. The
continuous onset and exponent-one law are properties of the finite
analytic Gaussian loss, not evidence of thermodynamic criticality.

In Landau language, \(\tau-1\) is the mass term for the order parameter.
At high \(T\) it is positive, and the disordered collapsed branch with
\(\mu(x)=0\) is stable. At low \(T\), this coefficient becomes negative,
the collapsed branch becomes unstable, and an active ordered branch
appears in which the latent coordinate emerges from the prior state as
an input-dependent collective coordinate with nonzero posterior mean
\(\mu(x)=\langle z\rangle\).

Both the raw posterior mean-square \(\overline{\mu(x)^2}\) and the
scale-invariant signal fraction \(M^2(T)\) measure the input-dependent
displacement of the posterior along a latent coordinate. An analogy in
disordered systems is the Edwards--Anderson order parameter for a
\(Z_2\) reflection symmetry \(z\to -z\): the data-averaged mean may
vanish, but the squared local mean remains nonzero.

\subsection{Scale-Invariant Order
Parameter}\label{scale-invariant-order-parameter}

The reason the signal fraction is the robust order parameter can be seen
from the decoder-eliminated reconstruction term:
\begin{equation}\label{eq-scale-invariant-reconstruction}
\frac{1}{2\sigma_\text{dec}^2}
\overline{
\left\langle
(x-w_\star z)^2
\right\rangle
}
=
\frac{\lambda}{2\sigma_\text{dec}^2}(1-M^2).
\end{equation}

The optimized reconstruction depends only on the signal fraction
\(M^2\), not on the absolute latent scale \(A^2\). Under a latent
rescaling, \begin{equation}\label{eq-auto-017}
\begin{aligned}
\mu(x)&\to s\mu(x),\\
\sigma_z^2(x)&\to s^2\sigma_z^2(x),\\
w&\to w/s.
\end{aligned}
\end{equation}

the raw second moments \(\overline{\mu^2}\) and \(\overline{\sigma^2}\)
rescale, and so does their sum \(A^2\), but \(M^2\) and the optimized
reconstruction do not.

In practice, however, the scale-fixing contribution is proportional to
\(T\), and the scale constraint gets weaker at low temperature, where
the empirical result may deviate from the one-mode canonical solution.
The raw observables \(\overline{\mu^2}\), \(\overline{\sigma_z^2}\), and
\(A^2\) can therefore drift from their canonical values, while the
scale-invariant collapse law for \(M^2\) remains stable.

The local coordinate \(\tau\) is useful because the isolated one-mode
collapse threshold is always \(\tau_c=1\). The global scan coordinate
used in the experiments is instead \(T=\beta\sigma_\text{dec}^2/V\). For
a quadratic (PCA) mode with eigenvalue \(\lambda\) from a dataset with
total variance \(V\), the same collapse threshold is therefore
\begin{equation}\label{eq-auto-018}
T_c=\frac{\beta_c\sigma_\text{dec}^2}{V}=\frac{\lambda}{V}.
\end{equation}

\subsection{Latent Scale and Auxiliary
Observables}\label{latent-scale-and-auxiliary-observables}

The dimensionless objective in Eq.
\eqref{eq-dimensionless-one-mode-loss} also shows explicitly where the
latent-scale condition comes from: \begin{equation}\label{eq-auto-019}
\frac{\partial \ell_\star}{\partial A^2}
=
\tau\left(1-\frac{1}{A^2}\right).
\end{equation}

For \(\tau>0\), exact stationarity with respect to the remaining latent
scale \(A^2\) thus fixes \begin{equation}\label{eq-auto-020}
A^2=\overline{\mu(x)^2}+\overline{\sigma_z^2(x)}=1.
\end{equation}

In this canonical one-mode scale, \(\overline{\mu(x)^2}=M^2\) and
\(\overline{\sigma_z^2(x)}=1-M^2\). Empirically, however, the raw
learned coordinate can have
\(A_k^2(T)=\overline{\mu_k(x)^2}+\overline{\sigma_k(x)^2}\) measurably
different from one, especially for low values of \(\tau\). This is why
we use \(M_k^2\) as the main observable order parameter, whereas the raw
\(\overline{\mu_k(x)^2}\) can deviate from the theoretical predictions.

In the canonical one-mode solution the same branch gives
\begin{equation}\label{eq-auto-021}
\begin{aligned}
\overline{\sigma_z^2(x)}
&=\tau=\frac{T}{T_c},\\
\overline{\log \sigma_z^2(x)}
&=\log \tau
=\log T-\log T_c,\\
R
&= \frac12 \log\frac{1}{\tau}
=\frac12 \log\frac{T_c}{T}.
\end{aligned}
\end{equation}

so the posterior log-variance is linear in \(\log T\) with slope \(+1\),
while \(R\) is linear in \(-\log T\) with slope \(1/2\). This canonical
branch is a constant-variance ansatz for the posterior noise; the linear
encoder used here is slightly more general because its log-variance head
is affine in the input. In the empirical analysis,
\(\overline{\sigma_k(x)^2}\), \(\overline{\log\sigma_k(x)^2}\), and
\(R_k\) are therefore compared as separately saved observables. Appendix
\ref{sec-latent-scale-diagnostics} shows the deviations in \(A_k^2(T)\)
and in the Jensen gap \(J_k(T)\).

The active-branch distortion is \begin{equation}\label{eq-auto-022}
D
= D_0\,\tau
=D_0\,\frac{T}{T_c}.
\end{equation}

Equivalently, after normalizing by the collapsed one-mode distortion
\(D_0\), the active-branch distortion is \(D/D_0=\tau=T/T_c\). The same
coefficient that controls the Landau sign change therefore also controls
the distortion recovered by keeping the mode. Here that coefficient is
the variance \(\lambda\) of the scalar data direction. At \(T=0\), the
single retained direction reconstructs that one centered scalar variable
exactly; at finite \(T\), the KL term leaves the residual distortion in
Eq. \eqref{eq-auto-022}. Appendix \ref{sec-spectrum-backprediction} uses
this one-mode law to predict the full \(T\)--distortion curves from the
PCA eigenspectrum alone.

The one-mode calculation also has a second use: it gives the reference
Landau form in Eq. \eqref{eq-one-mode-landau-expansion} for a single
isolated latent direction near collapse. In nonlinear VAEs, the same
expansion could be used only as a local stability guide once a branch
and basis have been chosen. If the decoder-eliminated reconstruction
term is expanded to quadratic order around that branch, the resulting
local utility operator plays the role that the data covariance plays in
the linear Gaussian problem.

In the linear case, however, PCA provides predictions to compare against
because it diagonalizes the quadratic fluctuation operator globally. We
turn to this calibration in the next section.

\section{PCA Calibration}\label{pca-calibration}

Let the centered data covariance be
\begin{equation}\label{eq-pca-covariance}
\begin{aligned}
C
&=
U\,\operatorname{diag}(\lambda_1,\ldots,\lambda_d)\,U^\top,\\
V
&=
\operatorname{tr}C
=\sum_j\lambda_j.
\end{aligned}
\end{equation}

Diagonalizing the centered covariance, equivalently performing the SVD
of the centered data matrix, gives the PCA basis \(U\). Rotating the
inputs and the linear decoder by \(U\) leaves the Gaussian
reconstruction loss unchanged. In this principal-component basis the
quadratic data-fit term is diagonal, and the stationary linear decoder
can be chosen to align its active columns with PCA eigendirections, up
to rotations within degenerate eigenspaces. The full loss is then a sum
of independent one-mode problems, one for each eigenvalue \(\lambda_k\),
plus unused collapsed directions.

The only remaining distinction is between the local one-mode coordinate
and the global scan coordinate. For the \(k\)-th eigendirection,
\begin{equation}\label{eq-pca-local-coordinate}
\begin{aligned}
\tau_k
&=
\frac{\beta\sigma_\text{dec}^2}{\lambda_k}\\
&=
\frac{T}{\lambda_k/V}.
\end{aligned}
\end{equation}

Thus the one-mode collapse threshold \(\tau_k=1\) becomes
\begin{equation}\label{eq-pca-normalized-onset}
T_k=\lambda_k/V.
\end{equation}

This is the precise sense in which the single-mode Landau branch becomes
a PCA spectrum: PCA supplies the orthogonal directions in which the
quadratic problem factorizes, while the one-mode solution tells us how
each direction collapses as the price per nat is varied.

The Landau coefficient, the collapse threshold, and the reconstruction
utility are therefore all set by the same quadratic scale. The exact
stationary solution gives the collapse behavior: the signal fraction
\(M_k^2\) collapses linearly as we approach the threshold from below.
The raw posterior mean-square is then
\(\overline{\mu_k(x)^2}=A_k^2M_k^2\). In the original likelihood
parameters the raw-\(\beta\) collapse thresholds are
\begin{equation}\label{eq-pca-beta-threshold}
\beta_{c,k} = \frac{\lambda_k}{\sigma_\text{dec}^2}.
\end{equation}

The utility of adding the \(k\)-th ranked mode to a truncated
reconstruction is governed by the same eigenvalue. When that
eigendirection is collapsed, its unreconstructed variance contributes
the one-mode baseline \(D_0\) in Eq. \eqref{eq-auto-016} to the
distortion. On the active branch, Eq. \eqref{eq-auto-022} reduces this
residual contribution to \(D_0\tau\); at zero price per nat, the mode
removes it entirely. For several well-separated modes, the same branch
law makes the sampled-latent distortion piecewise linear in \(T\)
between neighboring collapse thresholds, with a slope set by the number
of currently active modes; Appendix \ref{sec-spectrum-backprediction}
checks this spectrum-level prediction directly. In likelihood units, the
reduction in distortion is therefore \begin{equation}\label{eq-auto-023}
\Delta D_k = \frac{\lambda_k}{2\sigma_\text{dec}^2},
\end{equation}

and after rewriting in normalized squared-error units,
\begin{equation}\label{eq-pca-normalized-utility}
\Delta \tilde D_k
=
\frac{\Delta D_k}{V/(2\sigma_\text{dec}^2)}
=\frac{\lambda_k}{V}.
\end{equation}

Thus \begin{equation}\label{eq-pca-utility-onset-calibration}
\Delta\tilde D_k=\frac{\lambda_k}{V}=T_k.
\end{equation}

The calibration in Eq. \eqref{eq-pca-utility-onset-calibration} states
our main result directly: the collapse threshold, reconstruction
utility, and PCA explained-variance ratio are the same normalized
spectral weight for ranked mode \(k\). The local behavior of the order
parameters is then fixed by the one-mode laws above: the signal fraction
\(M_k^2\) is linear in the distance to collapse, the canonical
log-variance is linear in \(\log T\) with slope \(+1\), and the
canonical rate is linear in \(\log T\) with slope \(-1/2\).

This gives a direct empirical test of the utility--threshold duality. In
the WorldClim experiments, we check whether the collapse thresholds
extracted from the scan match the utilities measured by truncated
reconstructions, and compare truncated reconstruction with PCA
cumulative explained variance. Training and data-split details are
collected in Appendix \ref{sec-worldclim-experiment-details}.

Our WorldClim linear-Gaussian results are organized around three main
figures and two appendix figures.

\begin{figure}[!htbp]
\centering
\includegraphics[width=\linewidth]{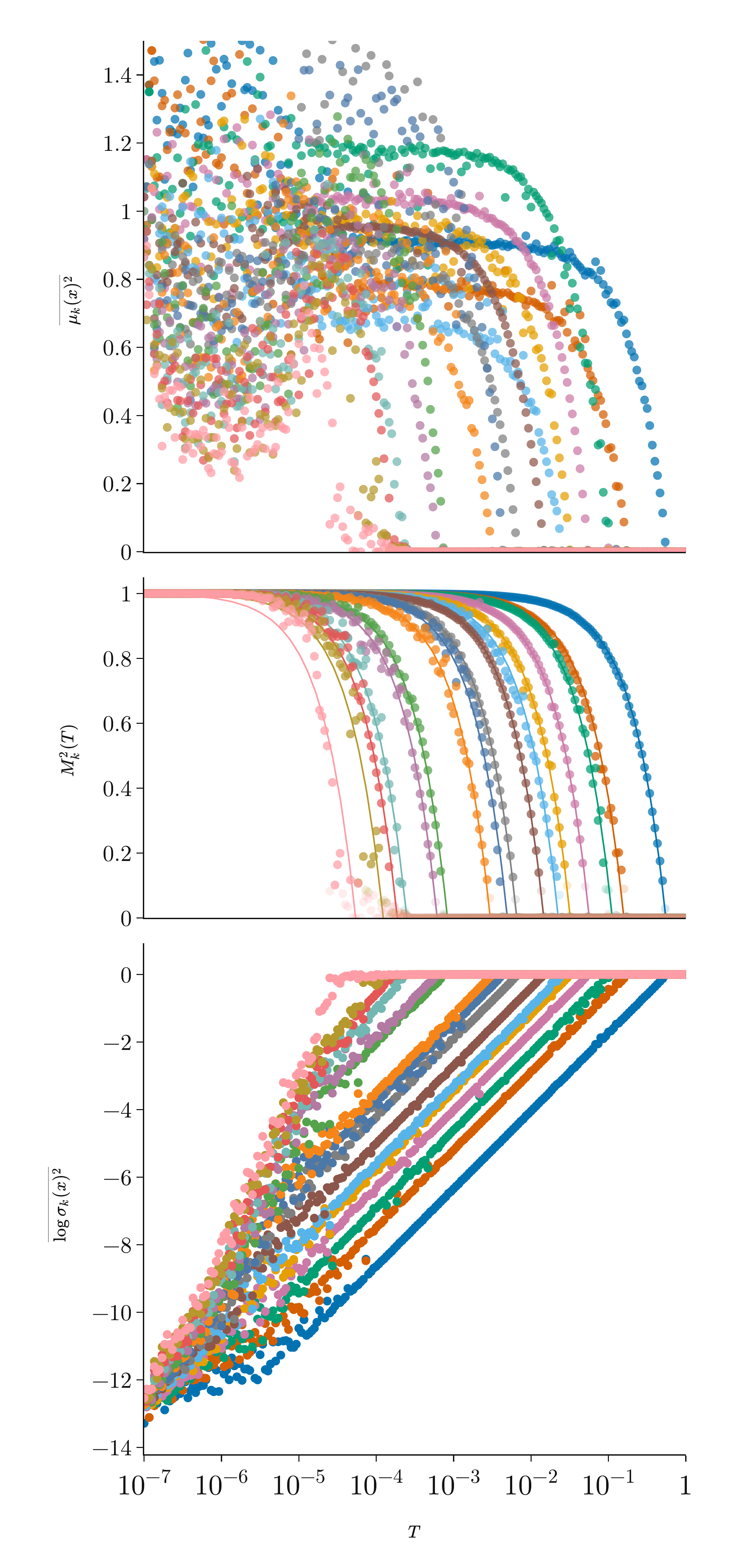}
\caption{\label{fig-worldclim-order-parameters}Order-parameter collapse scan.}
\end{figure}

\begin{figure*}[!htbp]
\centering
\includegraphics[width=\textwidth]{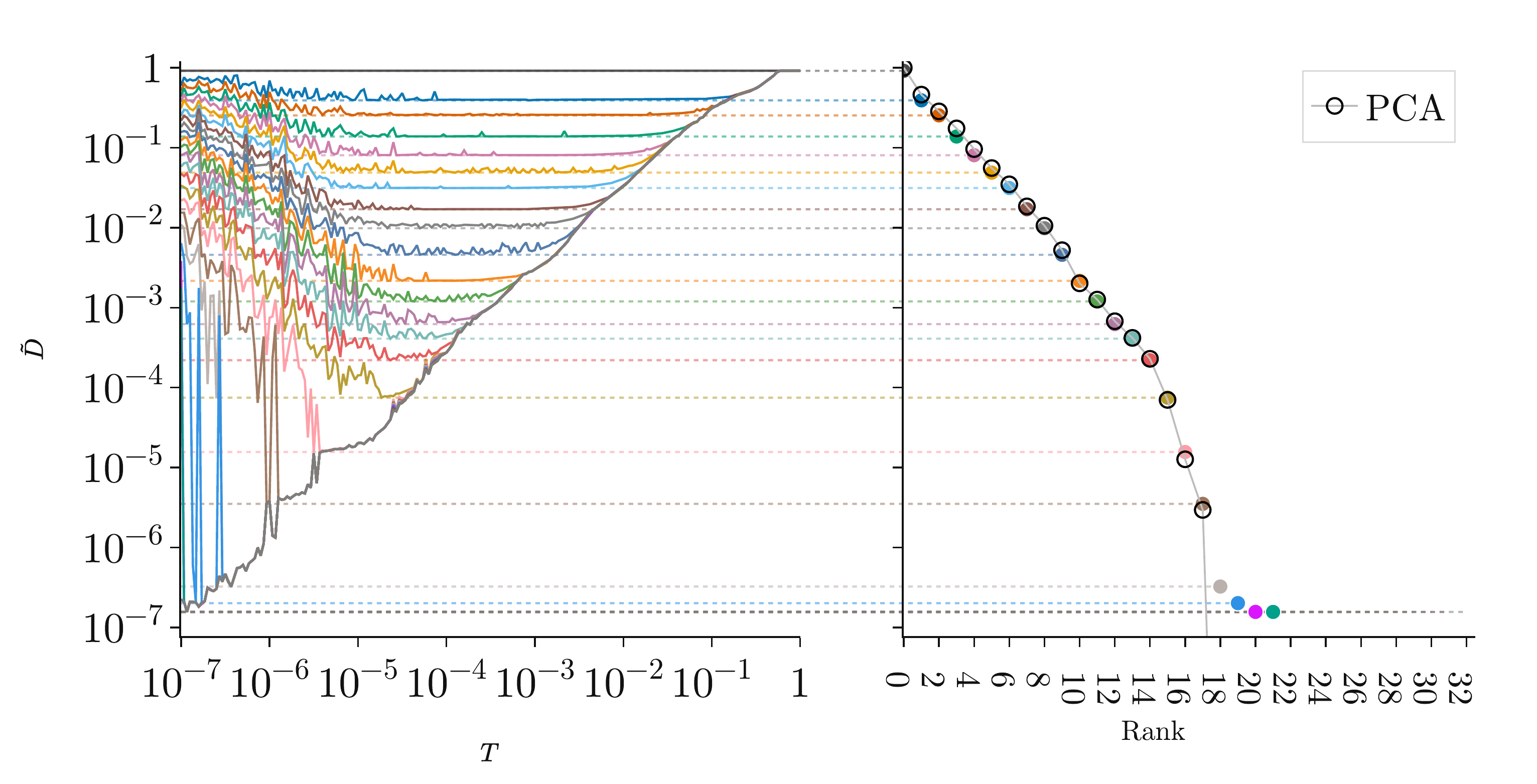}
\caption{\label{fig-worldclim-distortion-pruning}Truncated distortion and rank pruning.}
\end{figure*}

\begin{figure}[!htbp]
\centering
\includegraphics[width=\linewidth]{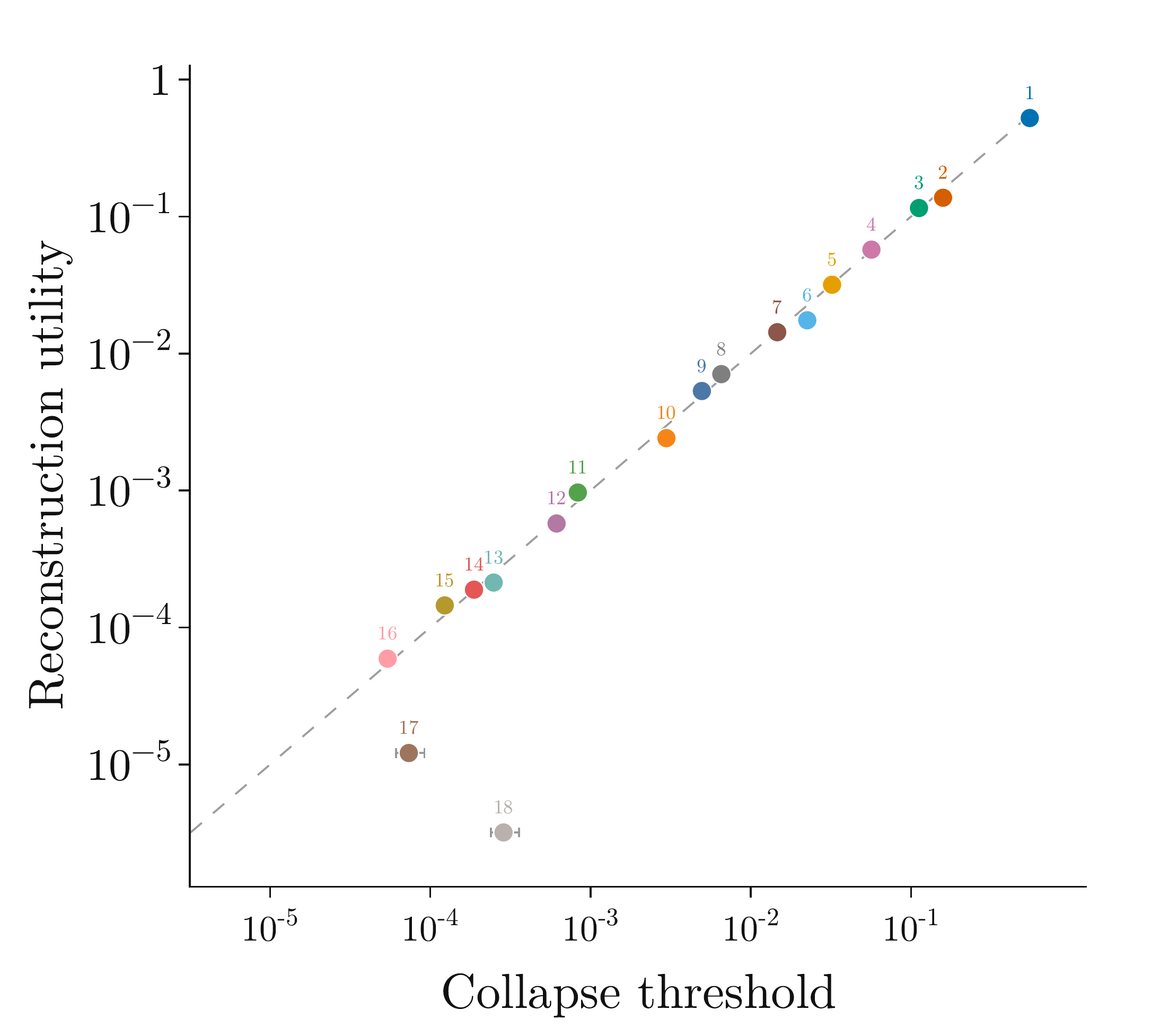}
\caption{\label{fig-worldclim-utility}Reconstruction utility vs collapse threshold.}
\end{figure}

\begin{enumerate}
\def\labelenumi{\arabic{enumi}.}
\tightlist
\item
  Fig.\ref{fig-worldclim-order-parameters} displays the raw posterior
  mean-square \(\overline{\mu_k(x)^2}\), the scale-invariant signal
  fraction \(M_k^2(T)\), and the posterior log-variance
  \(\overline{\log\sigma_k(x)^2}\) on a common scan axis. The collapse
  thresholds are extracted from the fixed-exponent fits in the \(M_k^2\)
  panel.
\item
  Fig.\ref{fig-worldclim-distortion-pruning} displays the truncated
  normalized distortion (using posterior means to remove sampling noise)
  as a function of the scan coordinate and as a function of retained
  rank, with the PCA reference included in the rank plot.
\item
  Fig.\ref{fig-worldclim-utility} displays reconstruction utility
  against the corresponding collapse thresholds, testing the
  utility--threshold duality itself.
\item
  Fig.\ref{fig-worldclim-spectrum-backprediction-appendix} in Appendix
  \ref{sec-spectrum-backprediction} checks whether the PCA eigenspectrum
  predicts the full \(T\)--distortion curves.
\item
  Fig.\ref{fig-worldclim-ak-jensen-appendix} in Appendix
  \ref{sec-latent-scale-diagnostics} diagnoses the behavior of the
  posterior scale \(A_k^2(T)\) and the posterior-variance Jensen gap
  \(J_k(T)\).
\end{enumerate}

Unless otherwise noted, points and curves in these figures are colored
by ranked mode.

The collapse scan is shown in Fig.\ref{fig-worldclim-order-parameters}.
The top panel shows the raw posterior mean-square
\(\overline{\mu_k(x)^2}\). The middle panel shows the scale-invariant
signal fraction \(M_k^2(T)\), with fixed-exponent collapse fits
overlaid. Least-squares fits are used on all points with \(M_k^2 > 0.1\)
to extract the collapse thresholds used in
Fig.\ref{fig-worldclim-utility}. The bottom panel shows the posterior
log-variance \(\overline{\log\sigma_k(x)^2}\). On the canonical one-mode
branch, this quantity is linear in \(\log T\) with slope \(+1\). The top
and bottom panels show visible noise for \(T \lesssim 10^{-5}\); the
middle panel hides most of this noise because \(M_k^2\) saturates at
one.

These order-parameter plots also address a practical interpretability
question: which latent should one inspect first? A priori, one could
answer this using almost any latent observable, or by directly selecting
coordinates according to reconstruction utility, for example through a
greedy-forward ranking. Fig.\ref{fig-worldclim-order-parameters} shows
why the raw posterior mean-square \(\overline{\mu_k(x)^2}\) is not a
good standalone answer: it displays level crossings due to latent-scale
drift (see Fig.\ref{fig-worldclim-ak-jensen-appendix}). In this linear
experiment, the posterior log-variance also gives clean ranked curves,
yet we use the signal fraction as the default ranking observable because
it is scale-invariant and directly tied to the one-mode loss.

Fig.\ref{fig-worldclim-distortion-pruning} makes the connection between
the collapse scan and the rank-resolved distortion curve explicit. The
left panel shows the truncated normalized distortion \(\tilde D_k(T)\)
for successive truncation ranks. Its horizontal guides mark the best
observed distortion level for each rank and help read plateaus and
rank-to-rank improvements. The right panel shows the corresponding best
distortion as a function of rank together with the PCA reference. This
cumulative view makes the spectral tail clear: the linear VAE normalized
distortion plateaus near \(10^{-7}\) by rank 18 onwards, while PCA rank
18 already reaches normalized distortion below \(10^{-12}\), and is
therefore outside the range of the log plot.

Appendix \ref{sec-spectrum-backprediction} gives a stricter check: the
PCA eigenspectrum predicts the full \(T\)--distortion curves, separately
for posterior means and for sampled latent variables, without fitting
the trained linear VAE utility spectrum.

Fig.\ref{fig-worldclim-utility} compares reconstruction utilities with
collapse thresholds extracted from the fixed-exponent \(M_k^2\) fits in
Fig.\ref{fig-worldclim-order-parameters}. Each point in
Fig.\ref{fig-worldclim-utility} represents one ranked latent coordinate.
The diagonal guide \(y=x\) highlights the utility--threshold duality: in
the ideal Gaussian baseline these quantities coincide exactly;
empirically, the points lie close to the diagonal. Utilities are
extracted directly from truncated reconstructions and remain fairly
stable through rank 18. The fitted collapse thresholds, however, depend
on the \(M_k^2\) fits and become inaccurate for rank 17 and higher.

\section{Spectral Regularization by Utility-Calibrated
Cutoff}\label{sec-spectral-regularization}

The normalized scan coordinate \(T\) acts as a spectral cutoff.
Coordinate \(k\) remains active if \(T < \Delta\tilde D_k\) and
collapses if \(T > \Delta\tilde D_k\), with the criterion applied mode
by mode rather than to a cumulative explained variance. Thus, in the
linear Gaussian setting, \(\beta\) regularization does not
indiscriminately shrink all latent coordinates. It selects latent
features one at a time: the VAE learns coordinates, ranks them by
reconstruction importance, and prunes those whose marginal utility falls
below the per-nat KL price.

This gives posterior collapse a concrete regularizing role. Some fitted
degrees of freedom may encode sampling noise, microscopic fluctuations,
or structure below the intended resolution of the model. If a noise
floor is known or estimated, \(T\) can be set at that scale, so
lower-utility coordinates collapse instead of being learned. In that
sense, the KL term is a spectral regularizer with a cutoff calibrated
here by reconstruction utility.

The analogy with field-theory regularization is limited but useful. In
physics, a cutoff is often clearest in a spectral variable rather than
directly in real space. Here the spectral variable is reconstruction
utility, the loss gain obtained by keeping a latent coordinate. Unlike
in a continuum-limit calculation, the cutoff here is not a regularizer
to be removed; it is a modeling choice that sets the resolution below
which variation is treated as residual detail. As dataset size grows or
measurement noise decreases, the appropriate cutoff may change, but it
remains explicit.

The posterior mean-square \(\overline{\mu_k(x)^2}\) plotted in
Fig.\ref{fig-worldclim-order-parameters} is an obvious first diagnostic,
but it is not the most robust one because it drifts under latent
rescalings. Several observables could in principle be used to rank
latents or extract collapse thresholds. The single-mode derivation above
selects the posterior signal fraction
\begin{equation}\label{eq-auto-024}
M_k^2 =
\frac{\overline{\mu_k(x)^2}}
{\overline{\mu_k(x)^2}+\overline{\sigma_k(x)^2}} .
\end{equation}

This observable is invariant under latent rescalings and is therefore
insensitive to latent-scale drift, making it a robust observable for
extracting collapse thresholds and ranking latent dimensions. Since
\(M_k^2\) is monotone in the aggregate posterior signal-to-noise ratio,
this also explains why SNR-based rankings can work in practice.

In Landau language, \(M_k^2\) is the order-parameter amplitude. For a
diagonal Gaussian posterior, Eq. (\ref{eq-auto-024}) is the fraction of
the aggregate posterior second moment carried by the input-dependent
mean rather than by posterior noise. It therefore measures the ordered,
data-dependent part of latent coordinate \(k\). Near collapse, the
reduced loss on this branch has the form
\begin{equation}\label{eq-auto-025}
\ell_k(M_k^2)
=
\ell_k(0)
+
\left(\frac{T}{T_k}-1\right)M_k^2
+
\frac{T}{2T_k}(M_k^2)^2
+
O\!\left((M_k^2)^3\right),
\end{equation}

so the collapsed state is stable for \(T>T_k\) and the active branch
appears for \(T<T_k\). In the VAE, this is not a phenomenological
assumption: it is the Taylor expansion of the exact one-mode loss after
the decoder has been minimized. The order parameter turns on
continuously and follows the exponent-one law obtained by minimizing
this analytic normal form.

This Landau picture should not be overextended. A recent treatment has
used criticality language for posterior collapse in VAEs
\citep{li2026posteriorcollapsephasetransition}. Our thermodynamic
analogy is deliberately limited; see Appendix
\ref{sec-relative-entropy-effective-temperature}. The free-energy-like
functional is the objective itself, not an approximation to a deeper
partition function. The latent variables are finite, model-defined
degrees of freedom, and their posterior averages are already contained
in the ELBO. No large-system limit is taken, no long-distance
correlations are inferred, and no intrinsic criticality of the data is
claimed.

Instead, the Landau form should be read as a stability analysis of the
crossover between collapsed and active states for each latent channel:
as \(T\) is lowered, an input-dependent posterior mean appears out of
the disordered prior state, with \(M_k^2\) measuring the strength of
this input dependence. In this language, the latent variable is the
candidate collective coordinate, while \(M_k^2\) measures whether that
coordinate remains input-dependent.

The scan therefore measures which collective coordinates remain active
at a given spectral cutoff, orders them by reconstruction utility, and
counts how many are needed to represent the data at a prescribed
accuracy. In the linear VAE these collective coordinates are PCA-like
linear combinations, and the cutoff is exactly calibrated by marginal
utility. In a nonlinear VAE the same diagnostic can rank learned
variables, but the utility scales and branch shapes must be measured
from the scan rather than predicted from PCA eigenvalues.

\section{Conclusions}\label{conclusions}

We have shown that posterior collapse in the linear Gaussian
\(\beta\)-VAE is an objective-level feature-selection mechanism, not
fundamentally a training failure. Increasing the KL weight raises the
price of keeping a latent coordinate input-dependent, set by \(\beta\)
in original units, by \(T = \beta \sigma_{\textrm dec}^2 / V\) in
normalized distortion units. The coordinates then collapse one by one
rather than shrinking uniformly, and in this solvable setting their
order is set exactly by marginal reconstruction utility.

The two spectra that define this selection rule are directly measurable.
For each latent dimension, the posterior signal fraction is the
scale-invariant order parameter for the collapse crossover, while
truncated reconstructions give the utility spectrum. Once \(\beta\) is
normalized in distortion units to become \(T\), the collapse-threshold
spectrum, the utility spectrum, and the normalized PCA spectrum
coincide. Since \(T\) selects coordinates through their collapse
thresholds, and those thresholds equal their marginal reconstruction
utilities, the cutoff acts directly on feature importance for
reconstruction.

This result clarifies what is distinctive about the VAE bottleneck. A
deterministic autoencoder can learn a low-dimensional code, and a sparse
autoencoder can encourage sparse activations, but SAE sparsity is
typically input-dependent: while each sample activates only part of the
dictionary, which elements are active can vary across samples. Neither
has the same global per-coordinate KL cost in the objective. In a VAE,
posterior collapse turns that cost into a dataset-level selection rule:
a coordinate remains input-dependent only as long as its reconstruction
utility justifies its information cost.

The linear result is also a calibrated null model. Beyond this baseline,
nonlinear decoders, non-Gaussian likelihoods, finite training and
optimization effects, and nearly degenerate regimes can rotate the
latent basis, mix modes, or deform the linear Landau collapse law. In
those settings the collapse spectrum will deviate from PCA; such
departures become measurable signals of nonlinear representation
learning rather than ambiguities of loss normalization.

As for the commonly-quoted ELBO choice \(\beta=1\), there is nothing
special about it in the Gaussian case: a change in \(\beta\) can be
absorbed into a corresponding change in the decoder-variance scale
\(\sigma_\text{dec}^2\), since only the normalized combination
\(T=\beta\sigma_\text{dec}^2/V\) affects the minimization. Even assuming
the common convention \(\sigma_\text{dec}^2=1\), the setting \(\beta=1\)
corresponds to \(T=1/V\), not to \(T=1\), and the same raw \(\beta\) can
therefore correspond to very different operating points---different
distortion, rate, and active rank---under different input or loss
normalizations.

The sensible procedure is therefore to choose a scale in normalized
units, i.e.~for \(T\) rather than for \(\beta\): given a target relative
tolerance, or an estimated normalized noise floor, \(\epsilon\), one
sets the cutoff to \(T\simeq\epsilon\) and then converts back to the raw
hyperparameter \begin{equation}\label{eq-beta-from-tolerance}
\beta \simeq \epsilon\,\frac{V}{\sigma_\text{dec}^2}.
\end{equation} For instance, setting \(\epsilon=0.01\) targets the
\(1\)\% utility scale: coordinates whose marginal normalized utility are
below that scale should collapse, while larger-utility coordinates
remain active. After training, the achieved distortion and active rank
are outputs of the model, not assumptions. If they are too high or too
low for the intended application, one can adjust \(T\) and retrain, or
scan locally around the initial value. This avoids a ``prune first, ask
questions later'' strategy: if one fixes a small latent budget from the
start, the achieved distortion is already constrained by the
architecture and cannot reveal what the full model could have reached at
the same operating point.

\section{Outlook}\label{outlook}

We have established the linear baseline for latent spectroscopy with
\(\beta\) scans. In the one setting where the answer is analytically
known, the scan recovers the PCA utility spectrum through posterior
collapse, rather than merely shrinking all coordinates at once. The
result is a ranked spectrum of latent-variable importance, measured by
reconstruction utilities and their matching collapse thresholds.

The scan also gives a way of testing low-dimensionality rather than
assume it. Instead of postulating that the data lie on a small set of
latent factors, it measures how many variables are needed to reach a
chosen reconstruction tolerance. A manifold-like description appears
only when the cumulative utility spectrum saturates rapidly; otherwise
the data require more modes, as in the linear case. The relevant object
is therefore not simply a latent space, but a set of latent variables
ranked by feature importance.

This suggests a limited analogy with thermodynamic and renormalization
group ideas. In physical systems, macroscopic behavior is often
controlled to a good approximation by a small number of emergent
variables or relevant operators, with additional variables entering only
when higher precision is required. The present work gives the simplest
VAE analogue of that idea: posterior collapse identifies which latent
coordinates are worth retaining at a given spectral cutoff, exactly
calibrated here by marginal utility as a feature importance.

Nonlinear models, coarse-graining, and dataset-size scaling provide the
natural next steps for testing whether such effective variables remain
stable, mix, or become irrelevant under changes of representation and
scale. In particular, nonlinear decoders may concentrate reconstruction
utility into fewer effective variables, producing a sparser active
spectrum at the same distortion tolerance. Latent spectroscopy provides
a direct way to measure this utility concentration and to test whether
the leading variables become more interpretable or disentangled at
particular regularizer strengths.

\appendix

\section{WorldClim Experiment
Details}\label{sec-worldclim-experiment-details}

The WorldClim experiments use the 19 bioclimatic variables from
WorldClim \citep{fick2017worldclim2} at 10 arc-minute resolution. We
keep grid cells that are valid in all 19 layers, giving 808,053
available samples. Inputs are standardized per feature using the
training split only.

For the 32-latent linear VAE run used in the manuscript figures, the
data are split with the spatial-block protocol used throughout these
experiments: equal-area blocks of side length 500 km, area-weighted
sampling, split seed 0, and the non-training blocks split between
validation and test. This gives 499,599 training samples, 153,897
validation samples, and 154,027 test samples, corresponding to 1952,
602, and 602 batches at batch size 256.

Each scan point is trained in equilibrium mode with an Adam optimizer,
learning rate \(3\times10^{-4}\), and a maximum budget of 500,000
optimizer updates. With 1952 training batches, this corresponds to about
256.1 epochs. The relative patience parameter is 0.1 of the update
budget, so early stopping can occur after 50,000 updates, or about 25.6
epochs, without a validation-loss improvement; otherwise training runs
to the update budget. In these experiments we follow the usual
\(\beta\)-VAE convention and set the decoder variance scale to one,
i.e.~\(\sigma_\text{dec}^2=1\). All reported scan coordinates are
converted back to \(T=\beta\sigma_\text{dec}^2/V\). The signal-fraction
threshold used in the scan diagnostics is \(M_k^2>0.1\).

\section{\texorpdfstring{Full \(T\)--Distortion
Curves}{Full T--Distortion Curves}}\label{sec-spectrum-backprediction}

In this Appendix, we compare the experimental \(T\)--distortion curve
with the analytic prediction using only the PCA spectrum as numerical
input. Let \(u_k=\lambda_k/\sum_j\lambda_j\) denote the normalized PCA
eigenvalue spectrum, sorted in decreasing order, and let \(d\) denote
the data dimension (\(d=19\) here). Starting from the rank-zero
distortion \(\tilde D=1\), the one-mode solution predicts how much
distortion each active mode removes:
\begin{equation}\label{eq-distortion-backprediction}
\begin{aligned}
\tilde D_z(T)
&=
1-
\sum_{k=1}^{d}
\max\!\left(u_k-T,0\right),\\
\tilde D_\mu(T)
&=
1-
\sum_{k=1}^{d}
\max\!\left(u_k-\frac{T^2}{u_k},0\right).
\end{aligned}
\end{equation}

The first line includes sampling noise from \(z\sim q(z\mid x)\) and
gives the already-known linear residual law. The second line is the
deterministic posterior-mean reconstruction. Its quadratic term comes
from the one-mode active law: the posterior mean suffers only a
shrinkage bias, so its residual distortion is bias-squared,
\(D_\mu/D_0=(1-M^2)^2=\tau^2\), whereas the sampled-latent distortion
keeps the posterior variance and therefore scales as \(D_z/D_0=\tau\).
Equivalently, each mode contributes the residual \(\min(u_k,T)\) in the
sampled case and \(\min(u_k,T^2/u_k)\) in the posterior-mean case. The
PCA spectrum and normalization are fitted on the training split, which
fixes the prediction before any other split is evaluated. In these fixed
training-set units, sufficiently large \(T\) leaves no mode active and
both expressions give the theoretical rank-zero baseline \(\tilde D=1\).
As \(T\) is lowered, modes become active in decreasing order of \(u_k\)
and progressively remove distortion. Because the full normalized
spectrum satisfies \(\sum_{k=1}^{d}u_k=1\), all \(d\) modes are active
at \(T\to0\) and the theoretical distortion vanishes.

\begin{figure}[!htbp]
\centering
\includegraphics[width=0.82\linewidth]{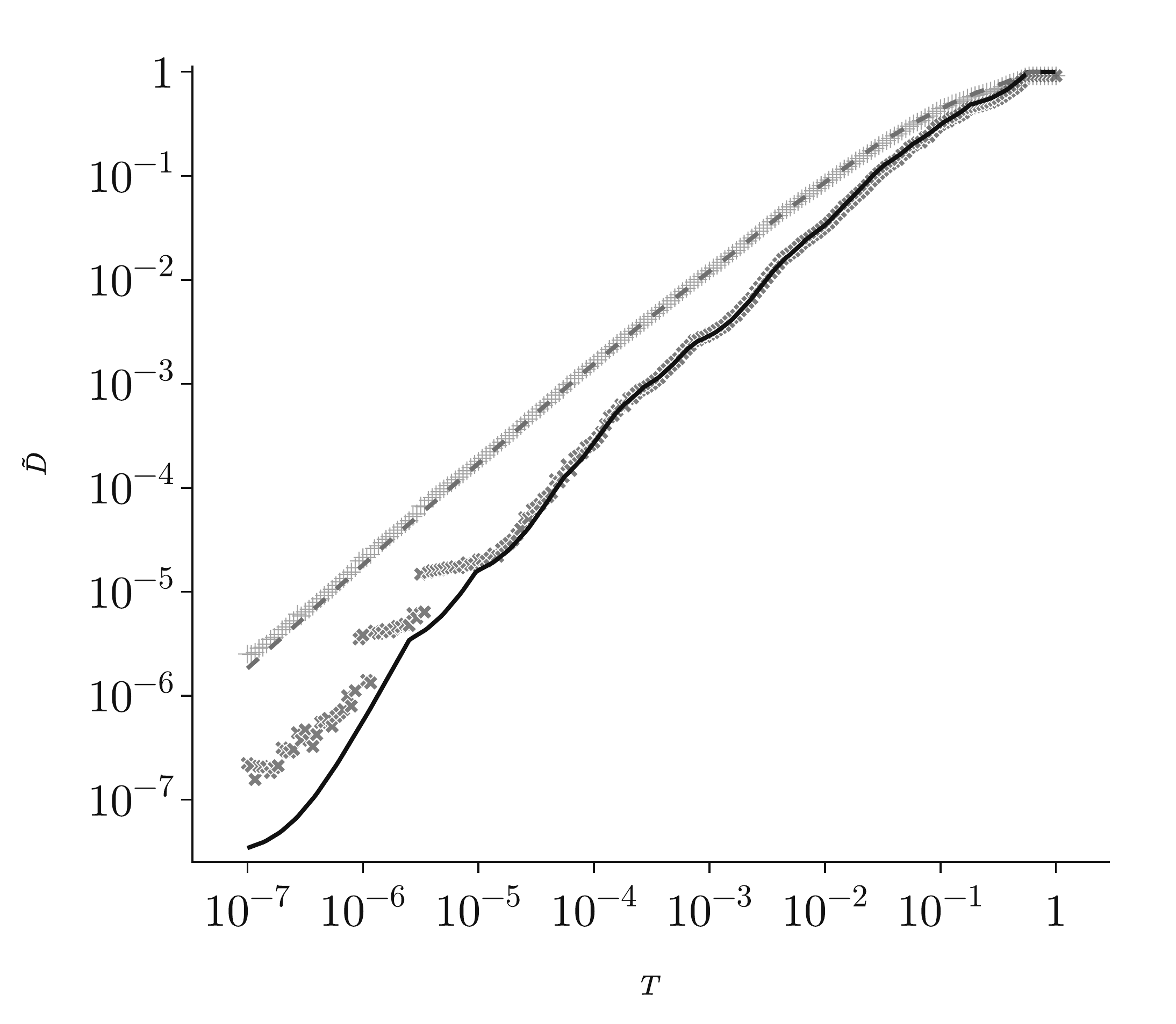}
\caption{\label{fig-worldclim-spectrum-backprediction-appendix}PCA prediction of the LinearVAE $T$--distortion curves. Symbols show held-out measurements for posterior means ($\times$) and sampled latents ($+$); solid and dashed lines show the corresponding parameter-free predictions from the normalized training-set PCA spectrum.}
\end{figure}

Fig.\ref{fig-worldclim-spectrum-backprediction-appendix} overlays these
predictions on the measured full-model \(T\)--distortion curves. No fit
is made to the full \(T\)--distortion curves or to the trained linear
VAE utility spectrum: the prediction uses only the full training-set PCA
eigenspectrum and normalization. It is therefore identical whichever
held-out split is plotted, whereas the measured linear VAE points depend
on that split and need not approach exactly \(\tilde D=1\) at rank zero.
Agreement holds over most of the resolved range. The visible exception
is the posterior-mean branch below \(T\simeq10^{-5}\), corresponding
roughly to rank 16 and beyond.

The linear-in-\(T\) statement applies to the sampled-latent distortion.
If the active set is fixed over an interval, with
\begin{equation}\label{eq-active-count-from-utilities}
n_d(T)=\#\{1\le k\le d:u_k>T\},
\end{equation} then Eq. \eqref{eq-distortion-backprediction} gives
\begin{equation}\label{eq-sampled-distortion-piecewise-linear}
\tilde D_z(T)
=
1-
\sum_{k\le d:u_k>T}u_k
+n_d(T)\,T .
\end{equation}

Thus, between two well-separated thresholds, \(\tilde D_z(T)\) is affine
in \(T\) with slope \(n_d(T)\). The posterior-mean reconstruction has
the corresponding active-set form
\begin{equation}\label{eq-mean-distortion-active-set}
\tilde D_\mu(T)
=
1-
\sum_{k\le d:u_k>T}u_k
+T^2\sum_{k\le d:u_k>T}\frac{1}{u_k},
\end{equation}

so it is quadratic in \(T\) within the same interval. This is why the
sampled and posterior-mean envelopes in
Fig.\ref{fig-worldclim-spectrum-backprediction-appendix} are not
expected to coincide, even though both are predicted by the same PCA
spectrum.

\section{Effective Temperature, Relative Entropy, and
Criticality}\label{sec-relative-entropy-effective-temperature}

In this Appendix we clarify the meaning and limitations of the
thermodynamic analogy used in the main text.

The regularization term in a VAE is the data-averaged KL divergence to
the latent prior, \begin{equation}\label{eq-auto-026}
R = \overline{\mathrm{KL}\!\left[q(z\mid x)\,\|\,p(z)\right]}.
\end{equation}

It is useful to distinguish carefully between absolute entropy, KL
divergence, and entropy-like quantities built from a reference
distribution.

On a finite discrete state space, Shannon entropy is
\begin{equation}\label{eq-auto-027}
H[q] = -\sum_i q_i \log q_i.
\end{equation}

If the reference distribution is uniform, \(u_i=1/N\), then
\begin{equation}\label{eq-auto-028}
\mathrm{KL}(q\|u) = \sum_i q_i \log \frac{q_i}{u_i} = - H[q] + \log N.
\end{equation}

Thus, in the discrete uniform-reference case, the KL divergence to the
uniform reference differs from negative absolute entropy only by an
additive constant. This is why one can often switch between ``entropy''
and ``relative entropy'' without much damage in finite statistical
physics.

For a non-uniform reference \(p\), however, one has
\begin{equation}\label{eq-auto-029}
\mathrm{KL}(q\|p) = -H[q] - \sum_i q_i \log p_i,
\end{equation}

so the relation is no longer a pure constant shift of \(H[q]\). This is
exactly the prior-relative situation relevant for VAEs.

In continuous space the situation is more subtle: differential entropy,
\begin{equation}\label{eq-auto-030}
h[q] = -\int q(z)\log q(z)\,dz,
\end{equation}

is not invariant under changes of coordinates and depends implicitly on
the reference measure. The invariant object is therefore not absolute
entropy but the KL divergence to a chosen reference density,
\begin{equation}\label{eq-auto-031}
\mathrm{KL}(q\|p) = \int q(z)\log\frac{q(z)}{p(z)}\,dz \ge 0.
\end{equation}

In the finite discrete case, the uniform reference is hidden in the
additive constant \(\log N\). On an unbounded continuous latent space,
no normalizable uniform reference exists, so an explicit reference
density is required to define a coordinate-invariant relative entropy;
in a VAE, the latent prior \(p(z)\) supplies that reference.

The sign convention also deserves care. In both information theory and
statistical physics, relative entropy normally denotes the nonnegative
KL divergence. To expose the free-energy structure without reversing
that terminology, define instead the entropy-like prior-relative
functional \begin{equation}\label{eq-auto-032}
S_p[q] \equiv -\mathrm{KL}(q\|p)\le 0.
\end{equation} Its maximum, \(S_p=0\), is reached at \(q=p\). For a VAE,
the data-averaged functional is \begin{equation}\label{eq-auto-033}
\overline{S_p}
\equiv
\overline{S_p[q(z\mid x)]}
=-R.
\end{equation} Because the KL divergence is nonnegative pointwise,
\(\overline{S_p}=0\) requires \(q(z\mid x)=p(z)\) for almost every
input: the coordinate is then fully collapsed and carries no input
information.

The VAE objective can therefore be written
\begin{equation}\label{eq-auto-034}
\mathcal L
=D+\beta R
=D-\beta\overline{S_p}.
\end{equation} This has the Helmholtz free-energy form \(F=E-TS\). In
the VAE convention, \(\beta\), or the normalized variable \(T\),
multiplies the entropy-like term and is therefore temperature-like. The
symbol \(\beta\) is inherited from \(\beta\)-VAEs and should not be
confused with the inverse temperature convention
\(\beta_{\mathrm{phys}}=1/T_{\mathrm{phys}}\).

Multiplying the complete loss by a positive constant, as one could have
done in Eq. \eqref{eq-normalized-loss}, leaves its exact minimizers
unchanged. Such a rescaling can still affect practical optimization
through finite step size, minibatch noise, optimizer state, and stopping
criteria.

The analogy is structural rather than literal. In canonical statistical
mechanics, minimizing the corresponding free-energy functional over
probability distributions produces the Gibbs ensemble; here, by
contrast, training minimizes the objective over encoder and decoder
parameters and their induced posterior distributions. The latent
coordinates are a finite set of internal variables, not local degrees of
freedom in a spatially extended system. Increasing the number of data
points does not provide a large-\(N\) limit of interacting degrees of
freedom, but only estimates the data average in the objective more
accurately. There is no inference of long-distance correlations and
therefore no correlation length whose divergence could be tested. The
collapse thresholds studied here are not thermodynamic critical
phenomena, but simply stability changes of the variational objective.

\section{Latent Scale Diagnostics}\label{sec-latent-scale-diagnostics}

Section III uses the canonical constant-variance one-mode branch. Our
linear encoder is slightly more flexible: its log-variance head is
affine in the input. We therefore keep \(\overline{\mu_k(x)^2}\),
\(\overline{\sigma_k(x)^2}\), \(\overline{\log\sigma_k(x)^2}\), and
\(R_k\) as separate saved observables, with no Jensen substitution of
\(\log\overline{\sigma_k(x)^2}\) for \(\overline{\log\sigma_k(x)^2}\).

The saved per-coordinate KL rate is not an independent collapse
observable: it is related to posterior log-variance via the latent-scale
drift \begin{equation}\label{eq-rate-logvar-scale-shift}
\begin{aligned}
R_k(T)
&=
-\frac12\overline{\log\sigma_k(x)^2}
+
\frac12\left[A_k^2(T)-1\right].
\end{aligned}
\end{equation}

\begin{figure}[!htbp]
\centering
\includegraphics[width=\linewidth]{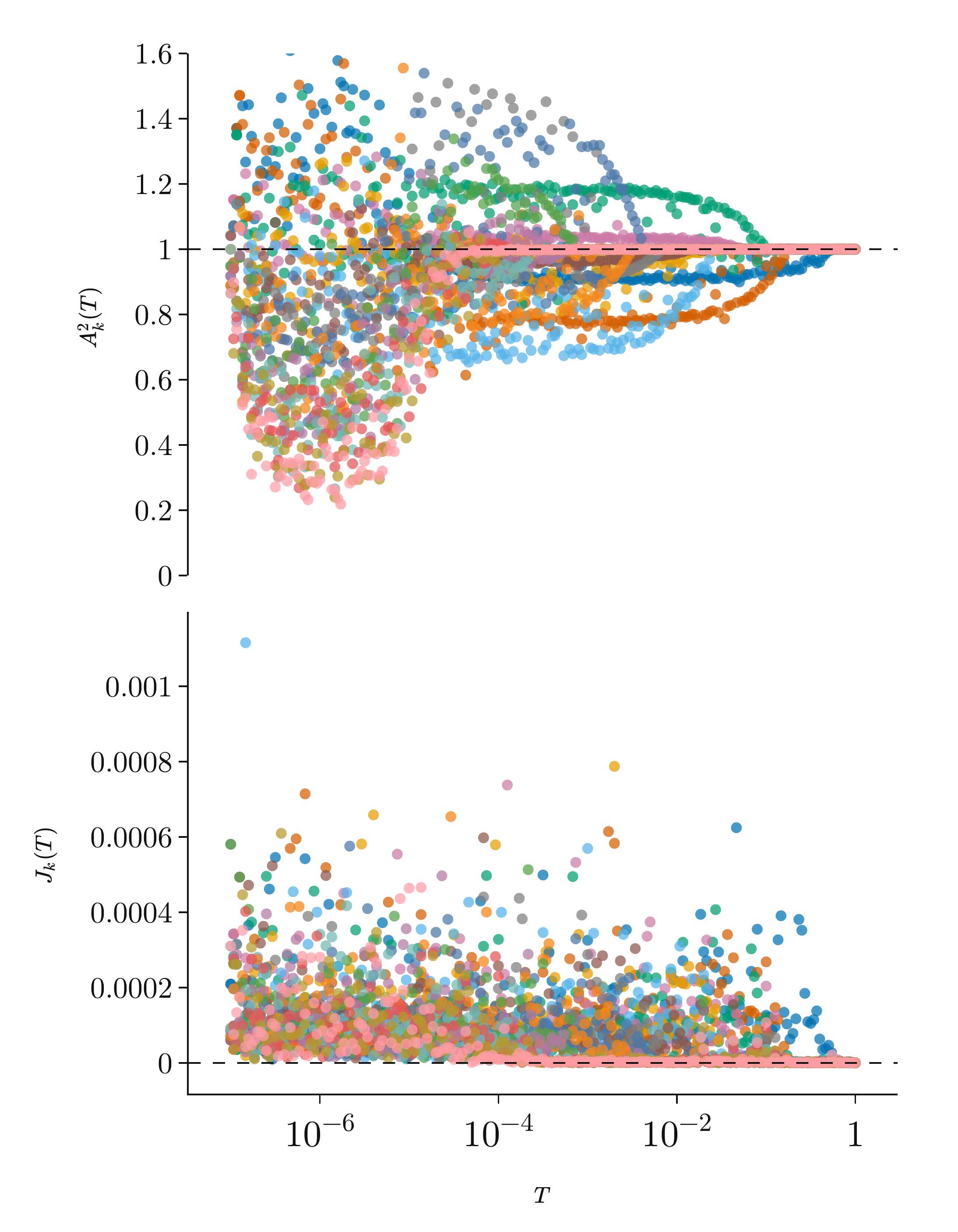}
\caption{\label{fig-worldclim-ak-jensen-appendix}WorldClim posterior-scale and Jensen-gap diagnostic for ranked latent coordinates.}
\end{figure}

Here \(A_k^2(T)=\overline{\mu_k(x)^2}+\overline{\sigma_k(x)^2}\) is the
corresponding aggregate second moment. Thus, if \(A_k^2=1\), the rate
and log-variance carry the same information in different units.
Departures of \(A_k^2(T)\) from one diagnose posterior-scale drift, not
a failure of the scale-invariant signal-fraction law. The
posterior-variance Jensen gap
\(J_k(T)=\log\overline{\sigma_k(x)^2}-\overline{\log\sigma_k(x)^2}\) is
a separate check of posterior-variance heterogeneity across samples.
Both diagnostics are shown for the WorldClim run in
Fig.\ref{fig-worldclim-ak-jensen-appendix}.

\bibliography{../references.bib}

\end{document}